# Analysis of Large-scale Traffic Dynamics using Non-negative Tensor Factorization

**Yufei HAN[1*] and Fabien MOUTARDE[2]**
1. TAO, INRIA Saclay, Rue 121, 91140 Orsay, France, yfhan.hust@gmail.com
2. CAOR, MINES-ParisTech, 60 Boulevard Saint-Michel, 75006 Paris, France

**Abstract** In this paper, we present our work on clustering and prediction of temporal dynamics of global congestion configurations in large-scale road networks. Instead of looking into temporal traffic state variation of individual links, or of small areas, we focus on spatial congestion configurations of the whole network. In our work, we aim at describing the typical temporal dynamic patterns of this network-level traffic state and achieving long-term prediction of the large-scale traffic dynamics, in a unified data-mining framework. To this end, we formulate this joint task using Non-negative Tensor Factorization, which has been shown to be a useful decomposition tools for multivariate data sequences. Clustering and prediction are performed based on the compact tensor factorization results. Experiments on large-scale simulated data illustrate the interest of our method with promising results for long-term forecast of traffic evolution.

## 1. Introduction

Floating-car data has become an essential source to determine the speed of traffic flows on the roads during recent years. This real-time traffic information is formed based on the collections of locations, speed and directions of vehicles through multiple kinds of mobile sensors, such as GPS-equipped vehicles, or drivers' mobile phones. Compared with loop detectors or surveillance cameras, floating-car data needs no additional hardware and generate estimation of variations of traffic flows in near real-time. Using floating-car data, traffic congestion can be detected and quantified measured, and travel times can be estimated efficiently. Due to these characteristics, many current research work and intelligent transportation systems make use of the floating-car data to simulate, model and analyze temporal dynamics of traffic flows.

Most published research on traffic data analysis focus on temporal dynamics of individual roads, or small groups of roads (in arterial network or freeways). Various data analysis tools are used to achieve this goal, which can be categorized either as model-driven [1][2] or data-driven approaches [3][4]. Generally, the model-driven approaches, like Cellular Automata and hydro-dynamics based models, construct generative parametric models using physical rules assumed for traffic flows. By calibrating their parameters with structural assumptions, the model-driven based method can simulate dynamic behaviors of traffic flows, which achieve modeling and prediction of traffic states at the same time. In contrast, the data-driven based methods relax the prior assumptions of traffic dynamics and extracts statistical descriptions of traffic flows efficiently using the methodologies originated from machine learning and signal processing. For example, Kalman filter and ARMA (Autoregressive Moving Average) [3][4], are used to track and predict temporal variations of traffic flows. Neural network improves the prediction performance by modeling with the non-linearity fluctuation of traffic flows with its multi-layer functional mapping structures. Notably, in [5], spatial correlations between local links are considered during temporal modeling of traffic flows using Markov Random Field. Following the simple criterion "Let the data speak for itself", the data-driven methods are more flexible than model-driven ones, and can track without prior assumptions the spatio-temporal dependencies of traffic states.

Interactions between the adjacent links are considered in both types of traffic data analysis. However, in a typical urban traffic scenario, congestion state of a region has high spatio-temporal correlations with its neighboring areas. Major improvement in estimating traffic dynamics therefore requires to model the dynamics of the whole network. Those facts motivate us to analyze the global dynamic patterns of large-scale networks. With a proper temporal model of the global traffic dynamics, we can estimate the spatial configurations of congestions in the network, which provides a global constraint in modeling the traffic behaviors of the whole network.

In our work, we propose to treat traffic states of all roads in a large-scale network as a whole, and unveil temporal dynamic patterns of the global traffic states. In a previous study [6], we had used matrix factorization to derive low-dimensional representation of global traffic state configuration, based on which we had achieved identification of typical spatial congestion patterns. This method only focused on spatial configuration of traffic states, while ignoring the temporal dependence between successive time periods. In this paper, we extend this idea and adopt a Non-negative Tensor Factorization (NTF) to describe spatio-temporal variations of traffic congestion in the network. We present the technical fundamentals about the NTF scheme in section 2. Clustering and long-term prediction of global traffic dynamics based on NTF are described respectively in sections 3 and 4. In section 5, we illustrate with experimental results in a large-scale synthetic traffic data the interest of our proposed algorithm. Section 6 concludes the whole paper.

## 2. Non-negative Tensor Factorization (NTF)

*2.1 Basics about tensor*

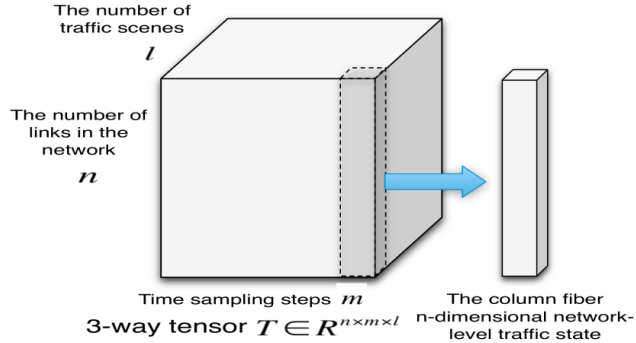

**Figure1. Structure of the traffic data 3-way tensor**

A tensor is defined as a multi-way array [7]. The order of a tensor, also known as its number of ways, is the number of indices necessary for labeling a component in the array. For example, a matrix, which is a 2-dimensional array, can be considered as a 2-way or $2^{nd}$ order tensor. Similarly, a vector and a scalar can be taken regarded as the first order and zero-th order tensors respectively. Due to the multi-way structure, tensor provides a coherent representation for the multivariate temporal sequence. As a matter of fact, tensors have been utilized popularly in video processing, chemometrics and psychometrics. Following this idea, we make use of a 3rd order tensor $T \in R^{n \times m \times l}$ to store temporal sequences of global traffic states, as shown in Figure 1. In our case, the numbers of elements along the three ways of $T$ are: $n$, $m$ and $l$, respectively corresponding to the number of links in the network, the number of time sampling steps in each daily traffic sequence and the number of traffic sequences in the historic data. Each entry $t_{i,j,k}(i=1,2..n, j=1,2...m, k=1,2...l)$ represents the traffic state of the i-th link captured at the j-th time sampling step of the k-th traffic sequence. As we can see on Figure 1, each column vector of the tensor $T$, defined as the column fiber, concatenates local traffic states of all $n$ links of the network. It contains the information about spatial configurations of traffic congestion, and we name it "network-level traffic state" hereafter. Each frontal slice of the tensor records the daily temporal variations of network-level traffic states, which is the main focus of this paper. According to the tensor structure, the three ways of $T$ illustrate separately the traffic behavioral patterns with respect to different links, time sampling steps and traffic sequences. Decoupling the three factors in large-scale traffic dynamics using tensor algebra will help us look into contribution of each factor in generating the variations of global traffic states.

*2.2 Non-negative Tensor Factorization*

Factorization of data into lower dimensional spaces provides a compact basis to describe the data and is sometimes also referred to as dimensionality reduction. Principle Component

Analysis (PCA) and Non-negative Matrix Factorization (NMF) [7] are the most popular data factorization methods. Experimental results show that the non-negativity constraint of NMF leads to a part-based decomposition of the data and improves clustering and classifying capabilities of multivariate data when compared to PCA. NTF takes NMF a step further by adding a dimension to the 2$^{nd}$ order tensor factorization (matrices) while inheriting its characteristics.

The basic idea of NTF is to decompose one tensor to be a canonical combination of matrices. The factorization procedure decouples the underlying factors along different ways of the tensor. To be more intuitive, we take NMF as an example. Note that NMF is a special case of tensor factorization applied in the 2nd order tensor. The formula of NMF is written in Eq.1:

$$A \approx \sum_{i=1}^{r}(u^i \circ v^i) = UV^T (U \geq 0, V \geq 0) \qquad (1)$$

where $A$ is a $n \times m$ matrix, $u^i$ and $v^i$ are the i-th columns of the $n \times r$ matrix $U$ and $m \times r$ matrix $V$, and $\circ$ is the outer product between vectors. Entries in both matrices $U$ and $V$ are constrained to be non-negative. Eq.1 presents a rank-r approximation to $A$. As presented in [7], treating each column of $A$ as a multivariate signal, the column space of $U$ and the row space of $V$ indicate correlated subsets of signal components and clustering membership of the multivariate signals in $A$ respectively. $U$ and $V$ represent physical characteristics of the signal and differences between signals. The matrix factorization decouples the two factors and represents them separately in the factorization matrices. By extending notions of NMF into 3-way tensor data, the NTF can be written as Eq.2 [7]:

$$T \approx \sum_{i=1}^{r} u^i \circ v^i \circ q^i \qquad u^i \geq 0, v^i \geq 0, q^i \geq 0 \qquad (2)$$

$u^i, v^i$ and $q^i$ are the i-th columns of three non-negative matrices: $U$ of size $n \times r$, $V$ of size $m \times r$, and $Q$ with size $l \times r$. Similar to NMF, the rows in $U$, $V$ and $Q$ refer to $r$-dimensional "fingerprints" or signatures of underlying traffic behavior patterns with respect to different links, time sampling steps and temporal traffic sequences respectively. Further analysis performed on the $r$-dimensional representations could unveil the spatial-temporal patterns of traffic behaviors in the network. In practice, NTF solutions can be obtained using an alternative updates algorithm [7], fixing all but one matrices among $U, V$ and $Q$, then updating the others using Karush-Kuhn-Tucker (KKT) condition.

## 3. Clustering of large-scale traffic dynamics

Based on Eq.2, fixing $U$ and $V$, we can represent each frontal slice of the 3-way tensor $T$ as a generalized linear basis expansion, as shown in Eq.3

$$T_k \approx \sum_{i=1}^{r} Q_{k,i} \times (u^i \circ v^i) \qquad (3)$$

$Q_{k,i}$ is the entry located at the k-th row and i-th column of $Q$, and each $(u^i \circ v^i)$ is a matrix with the same size as the frontal slice. Considering $\{u^i \circ v^i\}(i=1,2...r)$ as a basis of matrices, and each frontal slice $T_k$ as a matrix-form signal, Eq.3 illustrates a linear expansion of the matrix-form signal over the basis. $\{Q_{k,i}\}(i=1,2...r)$ is the r-dimensional vector of expansion coefficients on the basis. Note that we borrow the notion of multivariate signal expansion and extend them to the matrix-form objects. That provides an intuitive understanding of the tensor algebra. Due to the non-negativity constraint, the linear expansion in Eq.3 is a strictly additive superposition of the basis matrices to approximate $T_k$. Each matrix of the basis can be viewed as a component of the frontal slice. In our case, it leads to a part-based decomposition of the large-scale traffic dynamics. The expansion coefficients $\{Q_{k,i}\}(i=1,2...r)$ form a signature feature of large-scale traffic dynamic pattern in the corresponding k-th traffic sequence. Based on this property, we propose to perform clustering algorithm on the row space of $Q$, in order to find out the typical dynamic patterns of the network-level traffic state. For the clustering task, we choose the parameter $r$ in NTF to be 10. Higher $r$ provides more accurate approximation to the original tensor, while increasing complexity of the model. In the clustering procedure, we focus on classifying the dynamic patterns rather than reconstruction of the details. Therefore, we tend to choose smaller $r$ for clustering.

## 4. Prediction of large-scale traffic dynamics

The basic idea behind our work is a simple principle: given similar precedent traffic dynamic patterns, it is likely to observe similar subsequent temporal evolution of the network-level traffic state. In urban road networks, the topological structure of the network is stable and drivers' behaviors are historically consistent. Large historic data can cover most typical demand patterns of traffic resource, which provide comprehensive information about the global traffic dynamics. Based on this characteristic, we can achieve temporal prediction of the network-level traffic states by heuristically searching for the most similar temporal evolution in the historic observations.

We concatenate historic observations of traffic states into the tensor $T$, following the notions as Section 2. Assuming $M \in R^{n \times m}$ is a partially observed traffic sequence, only the first $m_1$ time sampling steps contain observations. Our work focuses on predicting temporal dynamics ranging from the $(m_1+1)$-th step until the end. Compared with one-step forecasts, the prediction task aims to cover a longer time period, thus named as "long-term prediction" in this paper.

Our proposed method makes use of the historic data heuristically. Firstly, we perform NTF on $T$ to learn the three factorization matrices $U$, $V$ and $Q$, as shown in Eq.1. Given enough historic data, $M$ can be considered as a point lying on the $r$-dimensional manifold spanned by the learned basis $\{u^i \circ v^i\}(i=1,2...r)$. Therefore, prediction of unobserved network-level traffic dynamics can be formalized as reconstruction of the missing columns in $M$ based on the learned manifold. The cost function used for our reconstruction procedure is

given in Eq.4:

$$J(Q^M) = \left\| M - \sum_{i=1}^{r} Q_i^M \times (u^i \circ v^i) \right\|_{Fro}^{Obs} + \lambda \sum_{j=1}^{K} s_{h_j} \left\| Q^M - Q_{(h_j,:)} \right\|_{L2} \quad (Q^M \geq 0) \qquad (4)$$

Here $Q^M$ is the $r$-dimensional estimated expansion coefficient of $M$ in the learned basis. $Q_i^M$ is the i-th entry of $Q^M$. We select K-nearest neighbors of $M$ among all frontal slices of $T$, indexed by $\{h_j\}(j=1,2...K)$. $s_{h_j}$ is the similarity measure evaluated between $M$ and the nearest neighboring frontal slice $T_{h_j}$, derived by calculating L2 distance between them with respect to the first $m_1$ observed columns. $\{Q_{(h_j,:)}\}(j=1,2..K)$ are the expansion coefficients of the K-nearest neighbors in the learned basis. $\|\|_{Fro}^{Obs}$ is the Frobenius norm calculated with respect to the first $m_1$ observed columns. The first term of the cost function evaluates reconstruction error between the observed ground truth of traffic dynamics and the NTF based approximation. The second term performs a locality preserving constraint on the estimated $Q^M$. By minimizing this term, the partially observed $M$ is regularized to be close to its nearest neighbors in the $r$-dimensional manifold spanned by the matrix-form basis, concatenating heuristic nearest neighbor information and suppressing artifacts of reconstruction. By adjusting the regularizing coefficient $\lambda$, we achieve a trade-off between the heuristic constraint of the unknown traffic dynamics and approximating accuracy of the known observations. The minimization procedure is performed by iteratively updating $Q^M$ using KKT condition [8]. Each step of the update can be written as Eq.5.

$$(Q_i^M)^{new} = (Q_i^M)^{old} \times \left( \frac{XWV^T + \lambda \sum_{j=1}^{K} s_{h_j} Q_{(h_j,:)}}{(Q_i^M)^{old} VWV^T + \lambda (Q_i^M)^{old} \sum_{j=1}^{K} s_{h_j}} \right)_i \qquad (5)$$

$(Q_i^M)^{new}$ and $(Q_i^M)^{old}$ are the values of the i-th entry in $Q^M$ before and after the current updating step. $X$ is a vector of $(n \times m)$ dimension containing elements of $M$ arranged in a column-wise order. $W$ is a diagonal square matrix with its side length equaling to $(n \times m)$. Its diagonal vector is a binary mask that gives 1 to the observed entries of $X$ and 0 to the missing ones. $V \in R^{(n \times m) \times r}$ is a matrix with each row storing the kronecker product of $u^i$ and $v^i$. The missing traffic states are then predicted by combining the estimated expansion coefficient $Q^M$ and the basis $\{u^i \circ v^i\}(i=1,2...r)$ following Eq.2. Different from the clustering task, we choose higher $r$ ($r=50$) to improve degree-of-freedom of the model to fit the data details.

# 5. Experimental Results
*5.1 Experimental settings and introduction of IAU-Paris database*

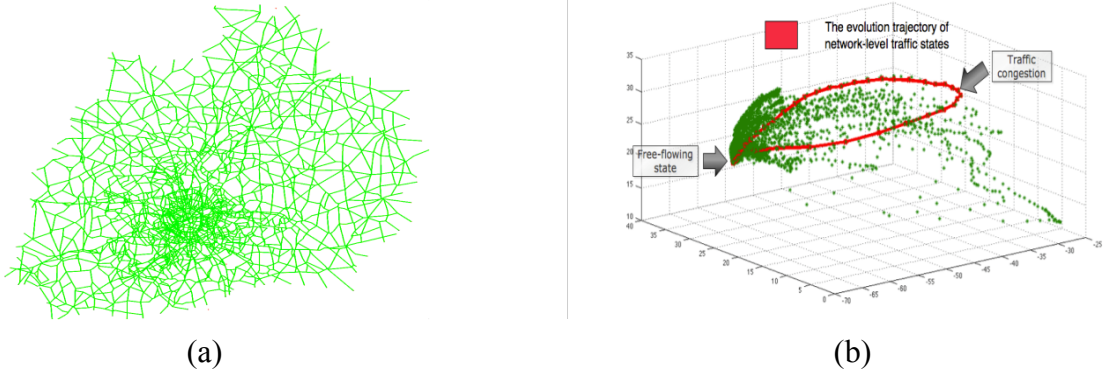

(a)            (b)

**Figure 2. (a) The large-scale roads network in IAU-Paris database.
(b) Temporal trajectory of network-level traffic states in 3D-PCA space.**

We test the proposed clustering and prediction method on a synthetic traffic data set, named as IAU-Paris database. This benchmark database contains 108 simulated traffic sequences of the large-scale Paris roads network, generated with Metropolis [9]. Metropolis is a planning software that is designed to model transportation systems. It contains a complete environment to handle dynamic simulations of daily traffic in one specific traffic network. The network is composed of 13627 links in Paris and its suburb region, as shown in the Figure 2(a). Each traffic sequence covers 8 consecutive hours of traffic data observations, including congestion in *morning* rush hours. Different traffic situations and evolutions are obtained by adding random events and fluctuations in the O-D matrix (Origin-Destination). There are 48 time sampling steps in each sequence, corresponding to 15-minutes bins over which the network traffic flow are aggregated. Traffic index [10] is used to represent traffic state in each link at a specific time, ranging in [0;1] interval. The smaller the traffic index is, the more congested is the corresponding link. To visualize large-scale traffic dynamics intuitively, we project the observations of the network-level traffic states into 3D-PCA space in Figure 2(b). Note that 3D-PCA is only used for visualization in the experiments. The three axis of the space correspond to the top three principal component axis. Each green point in the figure represents the 3D-PCA projection of one network-level traffic state observation. The points corresponding to the free-flowing states concentrate within a small region in the PCA space. The data points corresponding to moments when congestions occur in certain links are distributed sparsely and far from the region of the free-flowing state. Spatial configurations of network-level traffic states are similar if the whole network is almost free-flowing everywhere. On the contrary, congestion occurring at different parts of the network changes spatial patterns of traffic states in different ways, which introduces large variations into distributions of network-level traffic patterns. We use a red curve linking all network-level traffic states of the whole 48 time sampling steps in one simulated traffic sequence. The curve presents a circular and closed shaped trajectory in the 3D space. It indicates that the network-level traffic state evolves from morning free flowing, going through the peak-hour congestion, and then restores to be free flowing again at the end of the simulated sequence. Interestingly, the evolution of the network during increasing congestion follows a segment of trajectory that is totally different from the subsequent evolution of recovering to free-flow. This implies different temporal dynamics of traffic flows, and different intermediate traffic patterns, during increasing and decreasing of congestion.

*5.2 Clustering of the large-scale traffic dynamics*

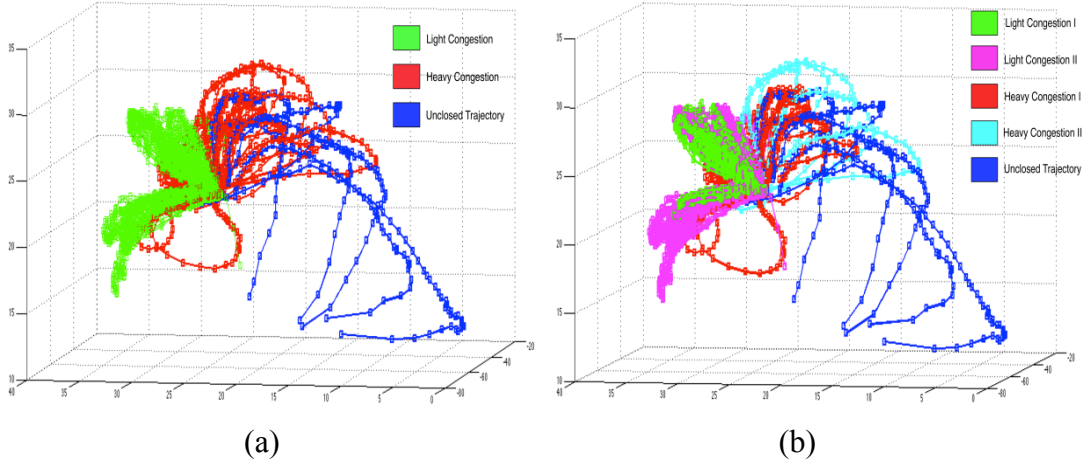

(a) (b)

**Figure 3. Clustering results of large-scale traffic dynamics**
**(a) The number of clusters is 3. (b) The number of clusters is 5.**

In the clustering task, we arrange all traffic index values in IAU-Paris database into a tensor $T \in R^{13627 \times 48 \times 108}$. After performing spectral clustering on the tensor factorization result, we obtain the clustering membership of each temporal sequence of the network-level traffic states. Different cluster represents different typical large-scale traffic dynamic patterns. We set the number of the derived clusters to be 3 and 5 respectively. Figure 3 illustrates the derived clustering structure. As shown in Figure 3(a), all 108 evolution trajectories are divided into three clusters. Temporal sequences in different clusters indicate traffic congestions of different degrees in the peak hour. The clusters labeled by the green and red legends represent the occurrences of light congestion and heavy congestion over the network, named as "Light Congestion" and "Heavy Congestion" respectively. Since heavy congestion introduces more diversity into the spatial configurations of traffic states, the temporal sequences in Heavy Congestion are distributed more diversely than those in Light Congestion, denoting more variations in large-scale traffic dynamics. The temporal sequences labeled by the blue legends don't have periodic temporal behaviors as the others, named as "Unclosed Trajectory". They correspond to occurrences of the overwhelming congestion at the beginning of the sequences. By increasing the number of clusters in Figure 3(b), we can get more detailed structure of the dynamic patterns. The Light Congestion cluster is divided into two sub-clusters, named as "Light Congestion I" and "Light Congestion II" in the figure. Trajectories in the two sub-clusters follow different orientations in the 3D-PCA visualization. It implies different temporal evolution patterns of global congestion configuration between the two sub-clusters. The Heavy Congestion cluster is split into two sub-clusters, as denoted by "Heavy Congestion I" and "Heavy Congestion II". Trajectories in Heavy Congestion II are distributed more diversely and sparsely than Heavy Congestion I, indicating occurrences of more severe traffic jam during the peaking hours. To provide more quantitative comparison between different dynamic patterns, we calculate the mean traffic index value over the whole network at each time sampling step along the trajectory. The sequence of the mean traffic index values represents an illustration of temporally varied traffic behaviors. For each of the five obtained clusters, we show the mean traffic index sequence corresponding to the center of the cluster, as seen in Figure 4. Particularly, the duration of congestion, the time period of the peak congestion and the mean index value of the peak congestion vary a lot among different average sequences, providing an intuitive knowledge about characteristics of different large-scale traffic dynamic patterns.

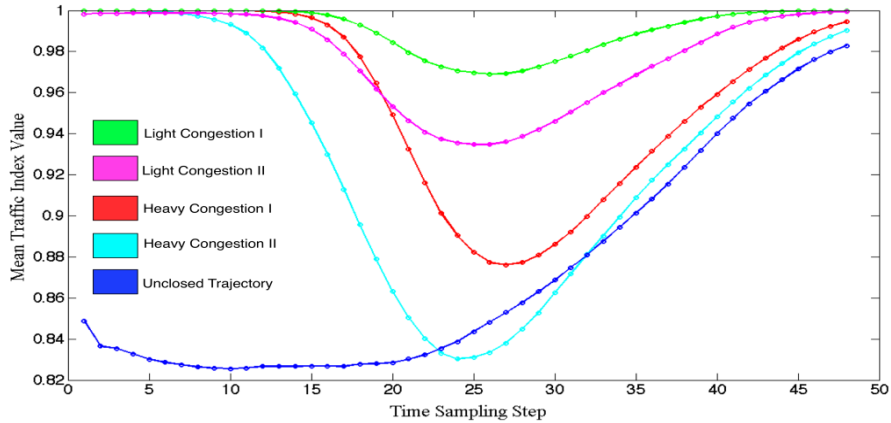

**Figure 4. Mean traffic index sequences for centroids of respective clusters**

*5.3 Experimental results of long-term prediction*

In the IAU-Paris database, about 1/3 of the 13627 links in the network stay free-flow most of the time, so should be ignored in metrics of congestion prediction. To be even more specific, we select only the most congested 3415 links for the prediction task. Besides, during the first 10 steps and last 5 steps, the network stays almost globally free-flowing. Therefore we only focus on the time period starting from the 11th steps to the 43th steps in the prediction task. Furthermore, in the left 31 time sampling steps, we choose the first 5 as the observed sub-sequence, which is the start of the peak hour in each traffic sequence. Temporal dynamics of the other 26 time steps are used for prediction. We select randomly 89 of the 108 simulated sequences as the historic observation records. The remaining 19 form the testing set. Given each tested sequence in the form of a frontal slice of a 3-way tensor structure $T^{test} \in R^{3415 \times 48 \times 19}$, we calculate the average L2 distance between the reconstructed entries and the corresponding ground truth in each frontal slice of $T^{test}$ to measure prediction accuracy. Larger average L2 distance means larger prediction error. The mean of the 19 distance measures corresponding to test sequences is used to evaluate the general prediction performances in the testing set, named as "General Prediction Error".

**Table 1. Prediction accuracies for our proposed method, with different settings of nearest neighbors**

| Number of nearest neighbors | General Prediction Error |
|---|---|
| 1 | 0.1789 |
| 3 | 0.1679 |
| 5 | 0.1695 |
| 7 | 0.1772 |
| 9 | 0.1835 |

**Table 2. Comparison of General Prediction Error for three methods**

| Method | General Prediction Error |
|---|---|
| Historic-Average | 0.2117 |
| Historic-NN (with K=3) | 0.1770 |
| The proposed NTF based method | 0.1679 |

We illustrate the variation of the prediction performances by increasing the number of nearest neighbors used in the reconstruction function (Eq.4) in Table 1. As we can find, in the testing set of IAU-Paris database, the number of nearest neighbors $K$ equaling to 3 is the best choice for predicting the unobserved traffic dynamics. By increasing the nearest neighbors from 1 to 3, the prediction performance increases gradually. On the contrary, further increase of the nearest neighbors leads to decline of the prediction accuracy. On one hand, more neighbors provide more heuristic information about the tendency of unobserved temporal dynamics. On the other hand, we should notice that more nearest neighbors do not assure better prediction. Neighboring historic data at the end of the KNN list might be deviated from the test traffic dynamics, which introduces noise into reconstruction and leads to the decline.

We compare the prediction performances of the proposed method with the other two baseline methods. The first one only calculates the average network-level traffic states of corresponding time steps in all sequences of historic data. They are used as the prediction results directly, named as "Historic-Average". In the second one, we make use of the heuristic information by calculating the historic average only on the K-nearest neighbors of each testing sequence, labeled by "Historic-NN". The comparison of prediction performances is listed in Table 2. Figure 5 illustrates the mean traffic index value of the total 26 time sampling steps involved in prediction in one testing sequence (from the 6$^{th}$ time sampling step to the $31^{st}$ step in the whole 36 steps). The mean traffic index values are derived from the ground truth and the prediction results of the three methods respectively. In Historic-NN and the NTF based method, we choose 3 nearest neighbors for prediction, according to results in Table.1. Historic-NN outperforms Historic-Average to a large extent. The Historic-Average doesn't make use of any heuristic information about temporal dynamics of traffic states. Purely average operation with respect to all historic data ignores the difference between traffic behaviors in different traffic sequences. Compared with Historic-NN, the proposed non-negative tensor factorization based method achieves further improvement. The proposed method is build by not only concatenating the heuristic neighboring information, but also extracting a representative model for large-scale traffic dynamics from the historic data through the tensor factorization, introducing more prior information about traffic dynamics

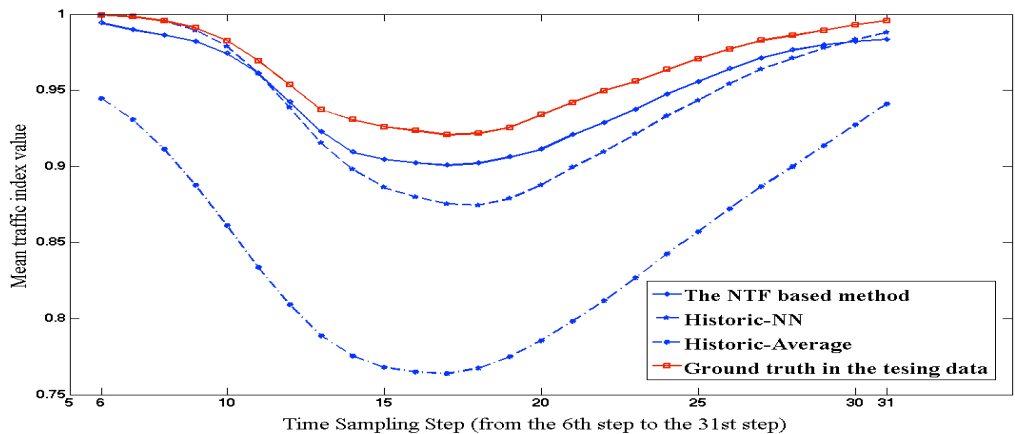

**Figure 5. Comparisons of the prediction performances**

## 6. Conclusions

We present a novel methodology for analyzing large-scale traffic dynamics, based on Non-negative Tensor Factorization. Our work contributes in three aspects. Firstly, we propose to use 3-way tensor as the basic level representation of traffic behaviors over the whole large-scale network. By arranging the data into a 3-way tensor, we aim to separate the link related, time interval related and daily traffic sequence related factors in large-scale traffic temporal behaviors. Non-negative tensor factorization scheme is then adopted to achieve this goal. Furthermore, by regarding the tensor factorization procedure as a matrix-form signal expansion, we derive a representative subspace based model for the global traffic dynamics of different traffic sequences. Using this compact model, we can conveniently perform statistical analysis on temporal evolutions of the global traffic states. Finally, we propose to combine the tensor factorization and KNN based heuristic information of the historic traffic data together in a tensor reconstruction framework. With the proposed method, we can reconstruct missing details of global traffic state configurations accurately.

Selection of the nearest neighbors plays a key role in our heuristic tensor reconstruction procedure, so most of our current improvement efforts relate to that part. It appears that optimal setting of the number of nearest neighbors depends on the design of the distance metric between traffic dynamics based on the partially observed entries. Also, the computation cost of selecting those nearest neighbors, currently using L2 distances, can become an issue on very large networks. We are therefore working on a more efficient way to select the nearest neighbors of large-scale traffic dynamics. Finally, our application is not limited to only long-term prediction, the proposed tensor reconstruction methodology could be further extended to estimation of missing observations of traffic states due to faults of sensors and noise in the GPS signals.

**Acknowledgement**

This work was supported by the grant ANR-08-SYSC-017 from the French National Research Agency. The author specially thanks Cyril Furtlehner and Jean-Marc Lasgouttes for providing advice and the benchmark database used in this article.

## References

1. Nagel, K., and Schreckenberg, M.: 'A cellular automaton model for freeway traffic', Journal of Physics,1992,2, pp. 2221–2229
2. Rakha, H.: 'Validation of Van Aerde's Simplified Steady-state Car-following and Traffic Stream Model', Transportation Letters: The International Journal of Transportation Research, 2009,1,(3), pp. 227-244
3. Wang, Y., and Papageorgiou, M.: 'Real-time freeway traffic state estimation based on extended kalman filter: a general approach', Transportation Research Part B, 2005, 39, pp.141–167
4. Statthopoulos, A., and Karlaftis, M.G.: 'A multivariate state space approach for urban traffic flow modeling and predicting', Transportation Research Part C, 2003,11, pp.121–135
5. Furtlehner, C., Lasgouttes, J., and De la Fortelle, A.: 'A belief propagation approach to traffic prediction using probe vehicles', Proc. 10th Int. Conf. Transportation Systems (ITSC), 2007, pp.1022–1027
6. Han,Y. And Moutarde, F.: 'Analysis of Network-level Traffic States based on Locality Preservative Non-negative Matrix Factorization', Proc.14th Int. Conf. Transportation Systems (ITSC), 2011, pp.501-506
7. Kolda, G.T. and Bader, B.W.: 'Tensor Decomposition and Applications', SIAM Review, 2009,vol.51, issue.3, pp.455-500
8. Lee, D.D., and Seung, H.S.: 'Algorithms for non-negative matrix factorization', Proc. 13th Neural Information Processing Systems (NIPS), Denver, USA, 2000, pp.556-562
9. Marchal,F.: 'Contribution to dynamic transportation models', Ph.D.dissertation, University of Cergy-Pontoise, 2001
10. Furtlehner,C., HAN,Y., Lasgouette, J., Martin,V., Marchal, F. and Moutarde,F., 'Spatial and Temporal Analysis of Traffic States on Large Scale Networks', Proc.13th Int. Conf. Transportation Systems (ITSC), 2010, pp.1215-1220